\documentclass{article}

   \usepackage[preprint,nonatbib]{neurips_2020}

\usepackage[utf8]{inputenc} % allow utf-8 input
\usepackage[T1]{fontenc}    % use 8-bit T1 fonts
\usepackage{hyperref}       % hyperlinks
\usepackage{url}            % simple URL typesetting
\usepackage{booktabs}       % professional-quality tables
\usepackage{amsfonts}       % blackboard math symbols
\usepackage{nicefrac}       % compact symbols for 1/2, etc.
\usepackage{microtype}      % microtypography
\usepackage{graphicx}
\usepackage{cite}
\usepackage{mathtools}

\usepackage{verbatim}
\usepackage{listings}
\usepackage{multirow}      
\usepackage{multicol}

\usepackage{algorithmic}

\newcommand{\ignore}[1]{}

\usepackage{listings}
\usepackage{xcolor}

\usepackage{hyperref}

\newcommand{\median}{\ensuremath{\mathrm{\textsc{Med}}}}

\def\BibTeX{{\rm B\kern-.05em{\sc i\kern-.025em b}\kern-.08em
    T\kern-.1667em\lower.7ex\hbox{E}\kern-.125emX}}

\usepackage[ruled,vlined]{algorithm2e}
 
\usepackage{algorithmic}
\usepackage{etoolbox}
\newcommand{\algorithmicdoinparallel}{\textbf{do in parallel}}
\makeatletter
\AtBeginEnvironment{algorithmic}{%
  \newcommand{\FORALLP}[2][default]{\ALC@it\algorithmicforall\ #2\ %
    \algorithmicdoinparallel\ALC@com{#1}\begin{ALC@for}}%
}
\makeatother

\usepackage{enumitem}

\setlist[enumerate]{itemsep=-1mm}

\usepackage{amsmath}

\title{BlockFLow: An Accountable and Privacy-Preserving Solution for Federated Learning}

\author{%
  Vaikkunth Mugunthan\thanks{ Denotes Equal Contribution} \\
  CSAIL\\
  Massachusetts Institute of Technology\\
  Cambridge, MA 02139 \\
  \texttt{vaik@mit.edu} \\
   \And
   Ravi Rahman\footnotemark[1] \\
   CSAIL\\
  Massachusetts Institute of Technology\\
  Cambridge, MA 02139 \\
  \texttt{r\_rahman@mit.edu} \\
    \And
   Lalana Kagal \\
   CSAIL\\
  Massachusetts Institute of Technology\\
  Cambridge, MA 02139 \\
  \texttt{lkagal@mit.edu} \\
}

\begin{document}

\maketitle

\begin{abstract}
%is a cyclical process where independent agents train, share, and merge machine learning models together, 
%Federated learning experiments typically incorporate differential privacy to protect the identity of those in the agents' datasets. 

Federated learning enables the development of a machine learning model among collaborating agents without requiring them to share their underlying data. However, malicious agents who train on random data, or worse, on datasets with the result classes inverted, can weaken the combined model. BlockFLow is an accountable federated learning system that is fully decentralized and privacy-preserving. Its primary goal is to reward agents proportional to the quality of their contribution while protecting the privacy of the underlying datasets and being resilient to malicious adversaries. Specifically, BlockFLow incorporates differential privacy, introduces a novel auditing mechanism for model contribution, and uses Ethereum smart contracts to incentivize good behavior. Unlike existing auditing and accountability methods for federated learning systems, our system does not require a centralized test dataset, sharing of datasets between the agents, or one or more trusted auditors; it is fully decentralized and resilient up to a 50\% collusion attack in a malicious trust model. When run on the public Ethereum blockchain, BlockFLow uses the results from the audit to reward parties with cryptocurrency based on the quality of their contribution. We evaluated BlockFLow on two datasets that offer classification tasks solvable via logistic regression models. Our results show that the resultant auditing scores reflect the quality of the honest agents' datasets. Moreover, the scores from dishonest agents are statistically lower than those from the honest agents. These results, along with the reasonable blockchain costs, demonstrate the effectiveness of BlockFLow as an accountable federated learning system.

%reward agents for their contributions toward the global model while preserving privacy and defending against malicious attacks.

% , who contribute models trained on random data or do not accurately evaluate others' models,

\end{abstract}

\section{Introduction}
Machine learning models benefit from large, diverse training datasets. For example, in healthcare, Electronic Health Records (EHR) provide a rich dataset for machine learning models that predict diseases \cite{zheng2017machine}. Though this would greatly benefit society at large, the sensitivity of the data and strict government regulations such as HIPAA \cite{annas2003hipaa} restrict how hospitals can share such data with other entities. As such, entities with sensitive datasets can only develop locally-optimal models. However, for robust, globally-optimal models,
that are highly generalizable, training needs to happen across organizational boundaries. Federated learning facilitates this by enabling the development of global models without sharing sensitive data \cite{konevcny2016federated}.  Individual agents train local models on their dataset and share their local models with a centralized agent, which computes and shares the averaged model. This process repeats for multiple rounds until convergence. \cite{mcmahan2017federated} demonstrated how federated learning leads to models that are stronger than any individual participant's local model. However, naive federated learning implementations are susceptible to several privacy and accountability threats.

%It enables distributed agents to train machine learning models without sharing underlying datasets. Federated learning is a multi-step process. 

%Let us consider the case where $N-1$ agents collude against an one agent in an $N$ agent federated learning setup. 

In an $N$ agent federated learning setup, $N-1$ colluding agents can compute the remaining agent's model. With the raw model, they can perform model inversion \cite{fredrikson2015model, wu2016methodology} and membership inference attacks \cite{shokri2017membership, nasr2019comprehensive, melis2019exploiting}. Differential privacy \cite{dwork2006calibrating}, the process of adding random noise to the model parameters \cite{bhowmick2018protection,mcmahan2017learning, bonawitz2017practical}, prevents these attacks. However, differential privacy does not protect against malicious agents that do not follow the protocol. For example, consider a logistic regression experiment where some agents invert their output labels, train, and submit a model based on this data. Such adversarial models can worsen the shared model \cite{kairouz2019advances}. Other agents would then use this worsened shared model in the next round, resulting in reduced overall accuracy \cite{li2019rsa,wu2019federated}. To defend against such attacks, agents must be held {\it accountable} for their contributions. Evaluating all contributions can prevent malicious contributions from being included in the shared model. Moreover rich evaluation can reward good contributors. Though it is trivial for a centralized agent to evaluate models such as by measuring accuracy on a secret (test) dataset, identifying a trusted, centralized agent is not always feasible (e.g. in international collaborations). Thus, a decentralized federated learning process, which is resilient to a minority of malicious agents, is ideal.

In addition, we propose the use of decentralized computational platforms, such as blockchains, for accountability. Blockchains combine an immutable ledger, currency, and with the Ethereum blockchain, a Turing-complete computational environment \cite{wood2014ethereum}. Programs that run in a blockchain's computational environment are known as ``smart contracts.'' All blockchain data is public, and the Ethereum consensus protocol ensures that smart contracts will be executed correctly. Thus, agents can trust that any data sent to the ledger will not be tampered with and that any blockchain currency (``cryptocurrency'') sent to smart contracts will only be released according to the contract's logic. With these guarantees, blockchains are appropriate to run federated learning accountability procedures and redistribute funds based on the results of these procedures.

BlockFLow is an accountable federated learning system that is fully decentralized and privacy-preserving. Its primary goal is to reward agents in relation to the quality of their individual contributions while protecting the privacy of the underlying datasets and being resilient to malicious adversaries. To accomplish these goals, BlockFLow requires no central trusted agent, enables agents to share differentially private models with each other and uses blockchain smart contracts to score each agent's contribution and provide monetary incentives for good behavior. Unlike existing auditing and accountability methods for federated learning systems, our system does not require a centralized test dataset, sharing of datasets between the agents, or one or more trusted auditors \cite{awan2019poster,kim2019efficient}. 

%Powered by Ethereum blockchain smart contracts, it is fully decentralized and resilient up to a 50\% collusion attack in a malicious trust model.

Our paper is organized as follows: In section \ref{sec:related_works}, we analyze other approaches to secure federated learning systems. Section \ref{sec:system} details the BlockFLow system design. Section \ref{sec:threat_analysis} elucidates BlockFLow's resistance against attacks. In section \ref{sec:evaluation}, we present experiments that validate BlockFLow's theoretical properties. In section \ref{sec:conclusion}, we summarize our contributions and discuss future work. We conclude with a discussion of the overall broader impacts of our work. 

%BlockFLow enables new types of federated learning experiments and discuss the overall broader implications of our work.

\section{Background and Related Work}
\label{sec:related_works}
Given the wide attack surface for federated learning, numerous techniques have been proposed to mitigate privacy and accountability threats. However, unlike our work,  previous approaches use a semi-honest trust model, require trusted agent(s), expose datasets, or do not defend against individual malicious agents.

%No existing accountability system preserves differential privacy, without requiring a centralized, trusted agent, and is resilient up to a 50\% malicious agent attack. 

\subsection{Differential Privacy}
Differential privacy \cite{dwork2010differential} is a mathematical framework that measures the risk of including an individual in a dataset. Differentially private methods can learn statistics of a dataset as a whole without learning sensitive information of individuals in the group. It guarantees with high probability that the released results do not reveal which individuals are in the underlying dataset. Formally, a randomized mechanism $\mathcal{M}$ satisfies $\epsilon$-differential privacy when there exists $\epsilon > 0$, such that $\text{Pr [}\mathcal{M}(D_1)\in S\text{]}\leq e^\epsilon \text{Pr [}\mathcal{M}(D_2)\in S\text{]}$ holds for every $S \subseteq $ Range($\mathcal{M}$) and for all datasets $D_1$ and $D_2$ differing on at most one element.

For any real-valued query function $t: \mathcal{D} \rightarrow \mathbb{R}$, where $\mathcal{D}$ denotes the set of all possible datasets, the global sensitivity $\Delta$ of $t$ is defined as $\Delta = \max_{\mathcal{D}_1\sim\mathcal{D}_2} |t(\mathcal{D}_1)-t(\mathcal{D}_2)|,$ for all $\mathcal{D}_1\in\mathcal{D}$ and $\mathcal{D}_2\in\mathcal{D}$.

The Laplacian Mechanism preserves $\epsilon$-differential privacy \cite{dwork2006calibrating}. Given $\Delta$ of the query function $t$, and the privacy loss parameter $\epsilon$, \textit{Laplacian mechanism} $\mathcal{M}$ uses random noise $X$ drawn from the symmetric Laplace distribution with scale $\lambda = \frac{\Delta}{\epsilon}$. For differentially private logistic regression, \cite{chaudhuri2009privacy} proposed $\Delta=\frac{2}{d*\alpha}$, where $d$ is the size of the participant's dataset and $\alpha$ is the logistic regularization parameter. Thus, adding noise drawn from $Laplacian(0, \frac{2}{d\alpha\epsilon})$ to logistic regression weights guarantees $\epsilon$-differential privacy.

Using differential privacy in federated learning can prevent collusion-based attacks and model inference attacks \cite{geyer2017differentially}. BlockFLow uses differential privacy to preserve the privacy of the agents participating in the federated learning system. 

%, it does not detect fallacious models

\subsection{Multi-Party Computation}
Multi-party computation (MPC) \cite{shamir1979share,Yao86,CCD87,goldreich1998secure,BGW88} guarantees that agents do not learn anything other than the final output. An MPC-based averaging mechanism can be used in federated learning for privacy, however, it is susceptible membership inference attacks and model inversion attacks. Additionally, MPC  requires a semi-honest threat model, which assumes that all agents follow the protocol correctly. BlockFLow offers a malicious agent threat model, where adversarial agents can be present, and uses differential privacy that prevents membership inference and model inversion attacks and thus provides stronger guarantees than MPC-based techniques.

%As such, MPC-based techniques are not appropriate when adversarial agents who may not follow the protocol are present.
%Membership inference attacks \cite{shokri2017membership, nasr2019comprehensive, melis2019exploiting} and model inversion attacks \cite{fredrikson2015model} on machine learning models can reveal the underlying raw data and other sensitive information of individuals participating in the dataset.

%For example, let us consider the scenario where $n-1$ agents collude against an honest participant in a $n$ participant federated environment and the agents want to compute the overall average. Once the agents compute the cumulative average, the dishonest agents can easily identify the private input of the non-colluding agent. Thus even if agents don’t share local data, it is still plausible for adversaries to learn and infer sensitive information from the model parameters that are being exchanged. 

\subsection{Blockchain-Based Accountability for Federated Learning}
Blockchain platforms have been used previously for accountable federated learning. 
\cite{kurtulmus2018trustless, harris2019decentralized} propose systems for evaluating models directly on the blockchain, via an initially-hidden but verifiable test dataset. While trust-less, these proposals require datasets to be publicly revealed on the blockchain and incur significant blockchain computational expenses for on-chain model evaluation. These proposals are only applicable for small models with public datasets.

Instead of a public on-chain evaluation, \cite{awan2019poster, kim2019efficient} propose using trusted aggregators and evaluators. These systems assume that all agents agree to trust centralized agent(s). While these proposals offer better privacy guarantees, for no data is publicly revealed, their weak trust model significantly limits their applicability.

Finally, \cite{chen2018machine} proposed a robust, trust-less, gradient-based validation scheme, where only differentially private gradient updates are revealed on a public blockchain. It only includes the gradient updates most similar to the average update. While both accountable and privacy preserving, error increases proportionally to the number of malicious agents. It also does not scale well to large models, for gradients are averaged on the public blockchain.

BlockFLow offers numerous advantages compared to these existing works. First, it is both privacy-preserving and trust-less: it does not require any secret test dataset, trusted set of agents, or the revealing of data or weights on a public blockchain. It also supports a robust threat model. So long as there are a minority of malicious agents, BlockFLow filters out all malicious agents and thus preserves shared model quality.

\section{Proposed System}
\label{sec:system}
% \begin{figure}
%     \centering
%     \includegraphics[width=0.5\textwidth]{BlockFLow-System-Architecture-Page-1.pdf}
%     \caption{BlockFLow System Architecture}
%     \label{fig:blockflow_system_architecture}
% \end{figure}

%BlockFLow is a decentralized system for privacy-preserving and accountable federated learning. 

BlockFLow is an accountable federated learning system that is fully decentralized and privacy-preserving. It leverages differential privacy to protect individual agents' datasets and an Ethereum blockchain smart contract to provide accountability. 

\begin{comment}
We provide reference implementations in the supplementary materials (appendix \ref{refImpl}) for both the Ethereum smart contract and the client code that interacts with the smart contract.
\end{comment}

\subsection{BlockFLow Client}
\label{sec:blockflow_client}
Each agent must run their own instance of the BlockFLow client, which handles all aspects of an agent's participation in a federated learning experiment. For every federated learning round, each BlockFLow client trains a local model, applies differential privacy by adding Laplacian noise to the model, shares its model with the other clients in the experiment, retrieves and evaluates other clients' models, reports the evaluation scores to the BlockFLow smart contract, and retrieves the overall scores and averages the clients' models.

The BlockFLow client interacts with the BlockFLow smart contract through an Ethereum client, which is required to communicate with the Ethereum network. Agents can run their own Ethereum client, such as Geth \cite{ethereum2017official}, or use a hosted Ethereum client, such as Infura \cite{infura}.

Since storing data on the Ethereum blockchain is expensive \cite{rimba2017comparing}, BlockFLow uses the InterPlanetary File System (IPFS) to share the differentially private models during each federated learning round. IPFS is peer to peer, and anyone can host any content. IPFS provides automatic tamper detection via cryptographic hashing \cite{benet2014ipfs}. Before uploading models to IPFS, clients encrypt their models using Elliptic Curve Diffie Hellman \cite{wang2008cryptanalysis} keys derived from the sender's and receiver's Ethereum accounts. Since IPFS is publicly accessible (and does not have any built-in authentication), encryption prevents external agents from accessing clients' models.

A detailed description of the client procedure is included in appendix \ref{algorithmClientProc}.

\subsection{Ethereum Smart Contract}
\label{sec:system:ethereum_contract}
The BlockFLow Ethereum smart contract implements our novel contribution scoring procedure. Each BlockFLow experiment must deploy its own instance of the BlockFLow smart contract on the Ethereum blockchain. In order to initialize and deploy the smart contract, the following needs to be specified: the number of federated learning rounds, a list of Ethereum accounts who are permitted to join the experiment, bond parameters to control the refund rewarding, and timing parameters that control the duration for each stage in the experiment.

At the beginning of an experiment, the smart contract requires that every client posts a bond (in Ether, the Ethereum cryptocurrency) to join the experiment. This bond is used to ensure that clients follow the protocol and to reward or penalize clients for their contributions.

Each federated learning round consists of five stages: \texttt{train}, \texttt{retrieve}, \texttt{evaluation commit}, \texttt{evaluation reveal}, and \texttt{compute score}. The smart contract enforces strict deadlines for each stage; clients who miss a deadline are eliminated from the smart contract and lose their posted bond.

Before the \texttt{train} deadline, clients must submit the IPFS address for their trained model to the smart contract. Before the \texttt{retrieve} deadline, each client reports to the smart contract the set of clients whose models were successfully validated. Next, the smart contract computes which clients were able to retrieve both the majority of the other clients' models and have other clients retrieve their model. Clients who fail this test are eliminated. Clients are required to evaluate all agents who pass this test. Before the \texttt{evaluation submit} deadline, clients must submit evaluation scores encrypted with unique salts.\footnote{A salt is random data used as an additional input to a one-way hash function to prevent dictionary attacks.} Then, before the \texttt{evaluation reveal} deadline, all clients must reveal their salts and decrypted evaluation scores to the smart contract. The smart contract verifies the salts and scores. Finally, in the \texttt{compute score} stage, the smart contract then executes the contribution scoring procedure as specified in section \ref{sec:system:contribution_scoring} to determine the overall scores for each client. The smart contract then uses these scores to perform a \textit{pro rata} partial refund of the bond among the clients, and clients use these scores to determine how heavily each model has to be weighted while averaging.

This process is repeated for every federated learning round. After the final round, the remaining bond is refunded among the clients.

\subsection{Contribution Scoring Procedure}
\label{sec:system:contribution_scoring}
BlockFLow requires each client to evaluate every other client's model during each federated learning round. Clients are encouraged to use their entire dataset to perform evaluations. For the set of models $W$ and evaluation function \texttt{eval} the following properties must hold:
\begin{align*}
    (\forall w, w' \in \{W\times W\}, w \ne w')& \quad \texttt{eval}(w) > \texttt{eval}(w') \iff \text{model } w \text{ is better than model } w' \\
    (\forall w \in W)& \quad \texttt{eval}(w) \in [0.0; 1.0] \\
\end{align*}
For classification problems, F1 or accuracy scores satisfy the above properties. The clients must agree on which formula to use.

Clients' overall scores reflect the minimum of a) the median score reported for their model (as determined by all clients) and b) the inverse of the maximum difference between one's reported score and the median score for each model. This procedure incentivizes clients to share high-quality models and perform high-quality evaluations. Submitting a low-quality model would result in a low model median score, and submitting inaccurate evaluations would result in a significant maximum difference from the median. Algorithm \ref{alg:scoring_procedure} provides a formal description of the scoring procedure.

\begin{algorithm}
\SetAlgoLined
    \begin{algorithmic}[]
        \FORALL{$\text{client pairs } \{a, k\} \in \{N \times N\}$}
            \STATE{$s_{a, k} \gets \texttt{eval}_a(k)$} \COMMENT{client $a$ performs off-chain evaluation of client $k$'s model using $a$'s own dataset and reports the score here. See section \ref{sec:system:contribution_scoring} for \texttt{eval} requirements.}
        \ENDFOR
        \FORALL{$\text{clients } k \in N$}
            \STATE{$m_k \gets \median{\{ s_{a, k} : \forall a \in N \}}$} \COMMENT{Compute the median model scores}
        \ENDFOR
        \FORALL{$\text{clients } k \in N$}
            \STATE{$m'_k \gets \frac{m_k}{\max{\{ m_k' : \forall k' \in N \}}}$} \COMMENT{Scale the median model scores to ensure a maximum of 1.0}
        \ENDFOR
        \FORALL{$\text{client pairs } \{a, k\} \in \{N \times N\}$}
            \STATE{$t_{a, k} \gets |s_{a, k} - m_k|$} \COMMENT{Compute the evaluation quality scores}
            \STATE{$t'_{a, k} \gets \max{(0, \frac{0.5 - t_{a, k}}{0.5 + t_{a, k}})}$} \COMMENT{Transform the evaluation quality scores}
        \ENDFOR
        \FORALL{$\text{clients } a \in N$}
            \STATE{$d_{a} \gets \min{\{t'_{a, k} : \forall k \in N\}}$ } \COMMENT{Compute the least accurate evaluation each client performed}
        \ENDFOR
        \FORALL{$\text{clients } a \in N$}
            \STATE{$d'_a \gets \frac{d_{a}}{\max{\{ d_{a'} : \forall a' \in N \}}}$} \COMMENT{Scale the least accurate evaluation scores}
        \ENDFOR
        \FORALL{$\text{clients } k \in N$}
            \STATE{$p_{k} \gets \min{(m'_k, d'_k)}$} \COMMENT{Compute the overall scores}
        \ENDFOR
        \RETURN{p}
    \end{algorithmic}
    \caption{\textbf{Contribution Scoring Procedure}. This procedure, except for the \texttt{eval} method, is implemented in the blockchain smart contract.}
    \label{alg:scoring_procedure}
\end{algorithm}

\section{Threat Analysis}
\label{sec:threat_analysis}
BlockFLow offers a 50\% malicious threat model. Specifically, in an experiment with $N$ agents, BlockFLow is resistant up to $M \in [0, \frac{N}{2})$ agents neglecting to follow the BlockFLow protocol for the experiment to maintain its integrity. We explore all known attacks on the BlockFLow system, on all components of the BlockFLow system, and offer how BlockFLow is resistant to such attacks.

\subsection{Privacy Threat Model}
\label{sec:threat_analysis:privacy}
By the definition of differential privacy, with extremely high probability, $N-1$ colluding agents cannot infer any information about the remaining agent \cite{bindschaedler2017achieving}. Note that $(\forall N \ge 3) \quad N-1 > \frac{N}{2}$.

\subsection{Ethereum Blockchain Threat Model}
\label{sec:threat_analysis:ethereum}
Public/private key cryptography and a proof-of-work consensus protocol secure the Ethereum Blockchain \cite{wood2014ethereum}. There are no feasible attacks on the Ethereum Network, without controlling 50\% of the computational power of the \textit{entire} Ethereum network. Such an attack is estimated to cost nearly \$150,000 USD per hour \cite{Crypto51} and has never been successful on the Ethereum mainnet \cite{coindesk}. As such, we do not consider it to be a feasible threat to BlockFLow.

As the Ethereum Blockchain is public and anonymous, clients could theoretically enroll multiple times in an experiment and thus have a disproportionate vote. This is analogous to ballot box stuffing \cite{lehoucq2002stuffing}. BlockFLow mitigates this attack through a restricted enrollment model, where only pre-approved Ethereum accounts can join an experiment. Through decentralized identity verification \cite{erc725,eips} or manual processes, agents can ensure that each other agent controls only one account. Such verification is beyond the scope of BlockFLow. So long as there are a majority of honest agents with only one enrolled account (and thus, one vote), malicious (colluding) agents could never control 50\% or more of the accounts in a BlockFLow experiment.

\subsection{IPFS and Data Sharing Threat Model}
\label{sec:threat_analysis:ipfs_data_sharing}
IPFS is immutable \cite{benet2014ipfs} and agents cannot change their model after submitting the cryptographic hash to the smart contract. The BlockFLow protocol requires each agent to report if it is able to load strictly more than ${\frac{N}{2}}$ models, and have strictly more than $\frac{N}{2}$ agents report the same for one's own model. The BlockFLow threat model guarantees that there are strictly more than ${\frac{N}{2}}$ honest agents, meaning that all honest agents alone would fulfill these conditions for each-other.

This scheme also guarantees that for all agents $a$ who meet these conditions, at least one different and honest agent $a' \ne a$ retrieved $a$'s model. In the worst-case scenario, of the strictly more than half of the agents who retrieved $a$'s model, strictly less than half are dishonest; the difference between these sets is never empty. Since IPFS allows anyone to share any content, one or more honest parties would share the model with all other agents. Thus, agents who are unable to retrieve a model directly from the source (e.g. due to firewall restrictions) would still be able to obtain all necessary models. Hence, this model sharing scheme satisfies the BlockFLow threat model.

\subsection{Contribution Scoring Procedure Threat Model}
\label{sec:threat_analysis:scoring}
There are several attacks possible on the contribution scoring procedure. First, we define \textit{malicious models} to have weights that are not reflective of a truthful dataset. For example, models trained on randomly generated data or on inverted output features are considered malicious. Naively averaging such models into a global model would likely harm the shared objective.

The BlockFLow contribution scoring procedure penalizes those who submit malicious models. All agents evaluate every other agent's model. The median score an agent's model receives handicaps that agent's overall score; lower scores result in less cryptocurrency received. In addition, when computing the global model at the end of each round, agents are encouraged to weight models in proportion to the overall scores received.

Next, we consider agents who collude during the training process to submit better models. For example, agents can secretly share raw data or models among $M < \frac{N}{2}$ colluding agents. All $M$ agents would then submit their best model. BlockFLow rewards agents who contribute strong models, and it is acceptable for multiple agents to submit identical models. All submitted models are evaluated independently and relative to each other. As such collusion is no different from having many agents with strong datasets, we do not consider training collusion to be an attack.

Finally, we explore attacks during evaluation. Through encryption and a commit-then-reveal protocol, the BlockFLow smart contract prevents agents from copying others' scores without collusion. Also, the Contribution Scoring Procedure maps any difference greater than 0.5 away from the model's median to a 0 score, as the a priori score is 0.5. Thus, agents are encouraged to evaluate honestly.

With collusion, a group of dishonest agents could submit the same realistic score, or even fabricate scores. Consider the extreme case where a minority subset of malicious agents report perfect 1.0 scores for their models and 0.0 scores for all others (e.g. models from honest agents). However, since there are strictly less than half malicious agents, and only the median model scores are used to determine one's overall score, the median score is guaranteed to be between the minimum and maximum scores reported by the honest agents. As any honest agent's score is a fair evaluation, it is impossible for colluding agents to materially affect the evaluation scores. 

Moreover, the fabrication of scores will only penalize those who attempt it. The contribution scoring procedure depicted in algorithm \ref{alg:scoring_procedure} limits one's overall score with the evaluation on which one was furthest away from the median. Specifically, any agent with an evaluation of more than 0.5 away from the median will receive an overall score of 0, and no share of the cryptocurrency pool. So long as the median scores are bounded by the range of the honest agents' scores, which our threat model guarantees, fabrication is never optimal.

\section{Evaluation}
\label{sec:evaluation}
We evaluated our system on the Adult Census Income (Adult) \cite{Dua:2019} and The Third International Knowledge Discovery and Data Mining Tools Competition (KDD) \cite{elkan2000results, Dua:2019} classification tasks. All discrete features are one-hot encoded, and continuous features were scaled globally and independently to be between 0 and 1. Two-thirds of the data was split in equal shares (unless otherwise noted) among the $N$ clients, without overlap, and clients reserved 20\% of their data for training validation. Clients used their entire datasets when evaluating others' models. The remaining one-third was reserved for testing. All experiments used logistic regression models with L2 regularization coefficient $\alpha=1.0$ and F1 scores as the \texttt{eval} function.

\begin{figure}[htbp]
    \centering
    \includegraphics[width=\textwidth]{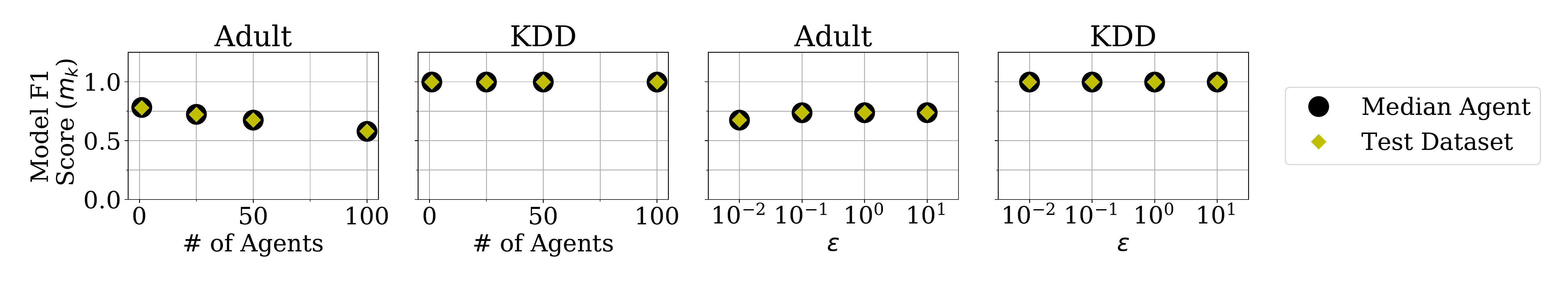}
    \caption{\textbf{Experiment 1: Measuring Model Quality} The above charts show the average of median agent F1 scores ($m_k$ from algorithm \ref{alg:scoring_procedure}) and average of reserved test dataset F1 scores. $\epsilon=0.01$ when varying $N \in \{1, 25, 50, 100 \text{ agents}\}$, and $N=50 \text{ agents}$ when varying the privacy parameter $\epsilon \in \{0.01, 0.1, 1, 10\}$. Across all evaluations, the average absolute difference is $< 0.67\%$.}
    \label{fig:exp1}
\end{figure}
In the first experiment, we validated that the median of the agents' F1 scores align with the one-third test dataset F1 scores, which we consider to represent absolute model quality. Agents did not train or evaluate on such test datasets. Figure \ref{fig:exp1} illustrates the similarity between these scores for all evaluated $(N, \epsilon)$ configurations. Thus, the median F1 score, even when individual agents have relatively small datasets, accurately estimates model quality.

\begin{figure}[htbp]
    \centering
    \includegraphics[width=\textwidth]{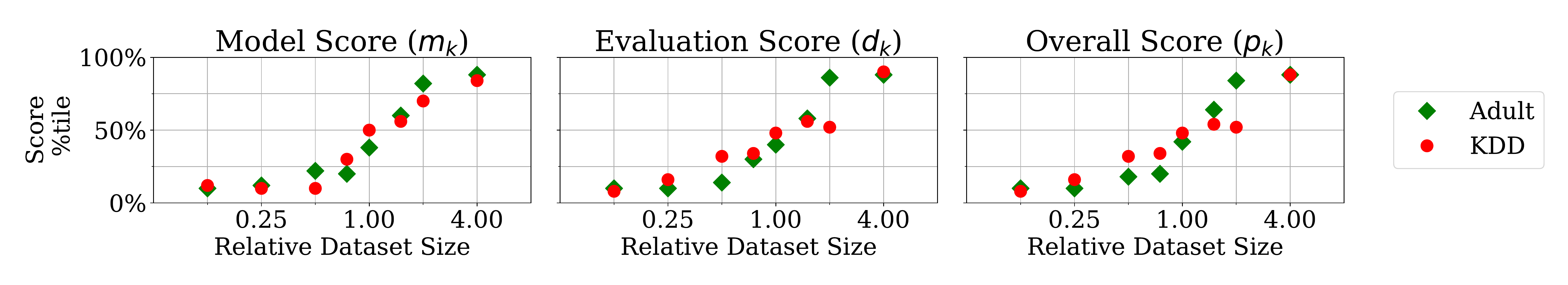}
    \caption{\textbf{Experiment 2: Rewarding Larger Datasets} Agent's model, evaluation, and overall scores ($m_k$, $d_k$, and respectively $p_k$ from algorithm \ref{alg:scoring_procedure}). Of $N=50$ agents, we varied the relative dataset size for 10 agents. $\epsilon=0.01$. For both datasets and all three scores, correlation $\rho(\log(\text{relative dataset size}), \text{score percentile}) > 0.9$. Hence, it is possible to identify and reward those with large datasets while maintaining dataset privacy and without sharing underlying data.}
    \label{fig:exp2}
\end{figure}
In the second experiment, we evaluated whether the contribution scoring procedure can reward those with higher-quality datasets, as higher-quality datasets should result in robust models and accurate evaluations. As shown in figure \ref{fig:exp2}, agents' model, evaluation, and overall scores were strongly correlated with their dataset size ($\rho > 0.9$ for all evaluations). As such, this scoring procedure can identify and reward (via cryptocurrency) those with larger datasets, without ever sharing the underlying data.

\begin{figure}[htbp]
    \centering
    \includegraphics[width=\textwidth]{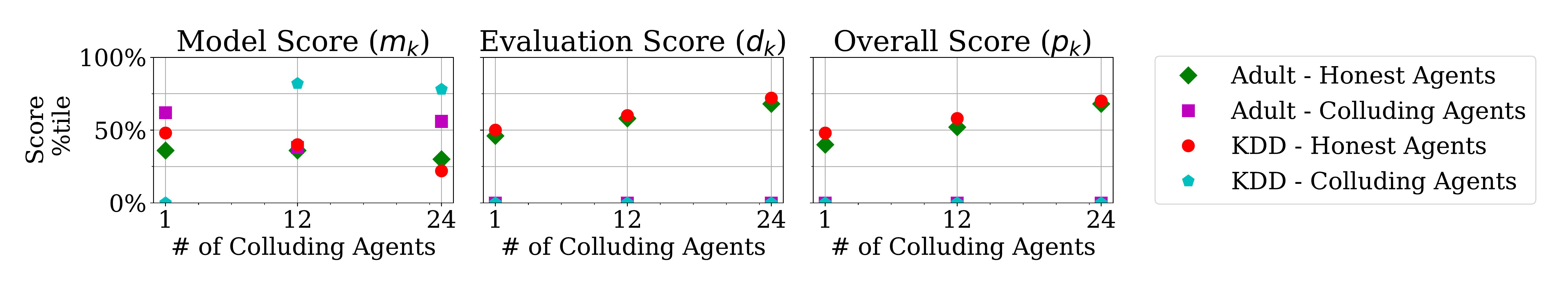}
    \caption{\textbf{Experiment 3: Discouraging Collusion} Average colluding and non-colluding agents' model, evaluation, and overall scores ($m_k$, $d_k$, and respectively $p_k$ from algorithm \ref{alg:scoring_procedure}). $N=50 \text{ agents}, \epsilon=0.01$. Agents colluded by rewarding other colluding agents perfect scores. While it appears that colluding agents obtained better model scores, note the statistically lower ($p < 10^{-31}$) overall scores for colluding agents compared to the honest agents. Hence, it is a sub-optimal strategy to fabricate evaluation scores.}
    \label{fig:exp3}
\end{figure}
In the third experiment, we examined evaluation collusion attacks. Specifically, consider the fabrication of evaluation scores where $M < \frac{N}{2}$ clients award perfect (1.0) scores to other colluding clients. Figure \ref{fig:exp3} illustrates how such collusion is reflected in the colluding agents' statistically lower evaluation and overall scores.

\begin{figure}
    \centering
    \includegraphics[width=\textwidth]{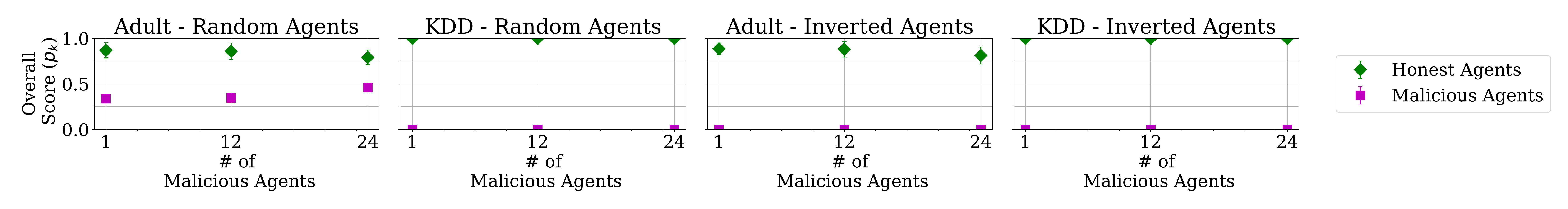}
    \caption{\textbf{Experiment 4: Punishing Malicious Agents} Average honest and malicious agent's overall score ($p_k$ from algorithm \ref{alg:scoring_procedure}). $N=50 \text{ agents}, \epsilon=0.01$. Random agents trained on random data; inverted agents trained out inverted output features. For all $N>1$ agent experiments, malicious agents' scores were statistically lower ($p < 10^{-22}$).}
    \label{fig:exp4}
\end{figure}

In the fourth experiment, we explored malicious training setups, where a minority subset $M < \frac{N}{2}$ of the agents submit models trained on \textit{random} or \textit{inverted} data. Independently sampling each feature with probabilities from the real dataset distribution formed datasets for random agents. Flipping the output labels created the inverted agents' datasets. Like honest clients, malicious clients split their datasets into 80\% train and 20\% validate portions. Figure \ref{fig:exp4} illustrates how the malicious agents scores were statistically lower than those of the honest agents.

Finally, we present an analysis of the blockchain costs of our system. Ethereum measures computational costs in terms of gas, which incorporates the amount of storage used and CPU instructions executed \cite{wood2014ethereum}. While the amount of gas is independent of market conditions, such as the price of gas or the exchange rate of Ethereum, it is directly proportional to real world costs. A linear regression ($R^2 > 0.99$) determined that per-agent gas consumption for $N$ agents and $R$ federated learning rounds is $$\text{gas}(N, R) = 31913N + 542045R + 477050NR$$ Such costs could be lowered by preferring off-chain computation of algorithm \ref{alg:scoring_procedure} and using the on-chain implementation only if any agent disagrees with the off-chain calculated scores.

Collectively, these experiments illustrate the robustness of the contribution scoring procedure and validate its theoretical properties discussed in section \ref{sec:threat_analysis:scoring}. The Laplacian Differential Privacy Method \cite{dwork2006calibrating}, Ethereum Blockchain \cite{wood2014ethereum}, and IPFS \cite{benet2014ipfs} and data sharing protocol are themselves resilient up to at least a 50\% attack. We have shown that the contribution scoring procedure is equally secure. As such, the entire BlockFLow system is resilient up to a 50\% malicious agent attack.

\section{Conclusion and Future Work}
\label{sec:conclusion}
By offering an accountable federated learning system that is fully decentralized and privacy-preserving, BlockFLow closes the trust and accountability gap with federated learning. No existing system simultaneously offers accountability guarantees, privacy preservation, and a fully trust-less, malicious agent threat model. This combination of features expands the reach for federated learning research, for example to international collaborations where agents need not know nor trust each-other but are rewarded for contributing high quality models.

BlockFLow can be extended to improve its performance. Specifically, its asymptotic costs, for $N$ agents and $R$ federated learning rounds, scale with $O(N^2 R)$, since for each round, each agent must evaluate all other agents' models. Instead of requiring all other agents' contributions, the smart contract could randomly select $Q << N$ agents to evaluate each agent's work. A sufficiently large $Q$, with high probability, would lead to accurate results and be resistant to $M < \frac{N}{2}$ malicious agents. This trade-off would greatly reduce gas consumption and increase scalability in large experiments.

\section*{Broader Impact}

This research is aimed at an often neglected aspect in federated machine learning - accountability of agents. Accountability is built into society through regulations and social norms but is rarely reflected in our technologies. Our research models how social accountability works through punishments and rewards and uses blockchain technologies to enforce them monetarily. We believe that this will be extremely useful for banks, hospitals and other organizations who want train machine learning models collaboratively but deal with very sensitive data and where erroneous training data could have serious consequences. For example, someone may try to deliberately sabotage the model (e.g. training on diabetes patients when in fact they are heart patients, or labeling babies as adults). BlockFLow can co-ordinate such experiments by ensuring that individuals' private information is not shared, detecting agents who are contributing poor models, and being resistant to 50\% collusion of malicious agents who try to sabotage the collaboration process. 

Federated learning ensures that models do not overfit and are highly generalizable. For example, health-care models trained on data collected from different hospitals located in different continents generalize better. Also, it is important to collect data from all demographics to build a fair classifier. BlockFLow leads to more unbiased and fair AI because it enables more generalizable models to be easily developed via federated learning.  

BLockFlow provides a payment mechanism to pay for high-quality (i.e. robust) models. Agents trying to cheat or induce bias will be caught with high probability, and the penalty cost imposed for cheating is more than the reward in expectation. This mechanism, along with BlockFLow's privacy guarantees and threat model, closes the trust and accountability gaps with federated learning. No existing system offers similarly strong properties. As such, BlockFLow enables new types of federated learning.

However, BlockFLow has some shortcomings that should be taken into consideration while designing the federation experiment. BlockFLow only compensates the data custodian, not the data owner. For example, patients are not directly compensated when a hospital uses patient data. BlockFLow cannot enforce that such compensation be made since differential privacy prevents anyone from knowing whether a particular individual was included in the training dataset. 

In addition, agents who lack computational resources or lack quality data would not be able to train a strong model and henceforth would receive less compensation. Though their contribution was valuable, such agents could end up losing money as their contribution was limited.  To address this, the scoring algorithm could be made binary (i.e.) all agents, whose models are of a certain quality, get the same reward rather than the better model getting a larger reward. In addition, agents with limited resources who cannot deposit a bond are prevented from participating in the federation system.

Lastly, if $M$ malicious agents $(M\geq\frac{N}{2})$ participate in the system, the system could fail and could result in a sub-optimal model. Along with this, the honest agents participating in the federation could lose their monetary benefits as well. 

%In spite of these drawbacks, the privacy of individuals is still preserved as each individuals' model is differentially private.

%Authors are required to include a statement of the broader impact of their work, including its ethical aspects and future societal consequences. 
%Authors should discuss both positive and negative outcomes, if any. For instance, authors should discuss a)  who may benefit from this research, b) who may be put at disadvantage from this research, c) what are the consequences of failure of the system, and d) whether the task/method leverages biases in the data. If authors believe this is not applicable to them, authors can simply state this.

%Use unnumbered first level headings for this section, which should go at the end of the paper. {\bf Note that this section does not count towards the eight pages of content that are allowed.}

\small 

\bibliographystyle{plain}
\bibliography{ref}
\newpage
\appendix
\section{BlockFLow Client Procedure}
\label{algorithmClientProc}
\begin{algorithm}
\SetAlgoLined
    \begin{algorithmic}[]
        \REQUIRE {agent $i \in N$; $i$ posted bond and enrolled in BlockFLow smart contract}
        \REPEAT
            \FORALL{$\text{agents } k \in N$}
                \IF{agent $k$ requests $w_{k'}^r$ for agent $k'$ and $w_{k'}^r$ is valid}
                    \STATE{encrypt $w_{k'}^r$ and share with agent $k$}
                \ENDIF
            \ENDFOR
        \UNTIL{$\text{forever, in background process}$}
        \FOR{$\text{round } r \gets 1 \text{ to } R$}
            \REPEAT
                \STATE $w^r_i \gets w^r_i - \eta \nabla g(w^r_i)$
            \UNTIL{deadline for round $r$}
            \STATE $w^r_i \gets w^r_i+Lap(0, \frac{2}{d_i\alpha\epsilon})$
            \STATE{$a_i^r \gets $ encrypt and upload model $w_i^r$ to IPFS and record the address}
            \STATE{record model IPFS address $a_i^r$ in BlockFLow smart contract}
            \FORALL{$\text{agents } k \in n$}
                \STATE{$a_k^r \gets$ get agent $k$ model's IPFS address from smart contract}
                \STATE{$w_k^r \gets \text{load and validate model } a_k^r$}
                \IF{$w_k^r \text{ is valid}$} \STATE{report $w_k^r$ as valid in BlockFLow smart contract} \ENDIF
            \ENDFOR
            \STATE{wait for data retrieval deadline}
            \FORALL{$\text{agents } k \in N$}
                \STATE{$v^r_k \gets $ count number of models agent $k$ marked as valid}
                \STATE{$z^r_k \gets $ count number of agents who marked agent $k$'s model as valid}
                \IF{$v^r_k \le \frac{N}{2} \text{ or } z^r_k \le \frac{N}{2}$}
                    \STATE{$N \gets N \setminus k$}
                \ELSE
                    \IF{agent $i$ does not have valid model $w_k^r$}
                        \FORALL{$\text{agents }k' \in N$}
                            \STATE{$w_k^r \gets $ request and retrieve model $k$ from agent $k'$} \COMMENT{See section \ref{sec:threat_analysis:ipfs_data_sharing} }
                        \ENDFOR
                    \ENDIF
                    \STATE{$s^r_{i, k} \gets \text{evaluate}_i(w_k^r)$}
                    \STATE{$b^r_{i, k} \gets \text{random encryption key}$}
                    \STATE{$s'^r_{i, k} \gets \text{encrypt}_{b^r_{i, k}}(w_k^r)$}
                    \STATE{report encrypted score $s'^r_{i, k}$ for agent $k$ to smart contract}
                \ENDIF
            \ENDFOR
            \STATE{wait for before encrypted score submission deadline}
            \FORALL{$\text{agents } k \in N$}
                \STATE{provide decryption key $b^r_{i, k}$ to provably reveal score $s^r_{i, k}$ to smart contract}
            \ENDFOR
            \STATE{wait for score decryption submission deadline}
            \STATE{$p \gets \text{ get scores from smart contract}$} \COMMENT{See Algorithm \ref{alg:scoring_procedure}}
            \STATE{$w_i^{r+1} \gets \frac{\sum_{k=1}^{n}(w^r_{k}*p_k)}{n * \sum_{k=1}^{n}(p_k)} $} \COMMENT{all agents should now have identical $w_i^{r+1}$}
        \ENDFOR\\
    \end{algorithmic}
    \caption{BlockFLow Client Procedure}
    \label{alg:client_procedure}
\end{algorithm}
\begin{comment}

\section{Reference Implementation}
\label{refImpl}
The reference implementation is available in the included \texttt{.zip} file.
\end{comment}
\end{document}